\documentclass[twoside,final,10pt,a4paper]{article}
\usepackage{graphicx,amsmath,natbib,rotating}

\begin{document}
\title{A Qualitative Dynamical Modelling Approach to
Capital Accumulation in Unregulated Fisheries}
\author{K. Eisenack$^a$, H. Welsch$^b$, J.P. Kropp$^a$\\
\\
$^a$Potsdam Institute for Climate Impact Research\\ 
P.O. Box 601203, 14412 Potsdam, Germany\\
Email: eisenack@pik-potsdam.de, kropp@pik-potsdam.de\\
\\
$^b$University of Oldenburg,
Dept. of Economics\\ P.O. Box 2503, 26111 Oldenburg, Germany\\
Email: welsch@uni-oldenburg.de}
\date{}
\maketitle

\begin{abstract}
Capital accumulation has been a major issue in fisheries economics
over the last two decades, whereby the interaction of the fish and
capital stocks were of particular interest. 
Because bio-economic systems are intrinsically  complex,
previous efforts in this field have relied on a variety
of simplifying assumptions. The model presented here
relaxes some of these simplifications.
Problems of tractability are surmounted
by using the methodology of qualitative differential equations (QDE). 
The theory of QDEs
takes into account that scientific knowledge about particular fisheries is
usually limited, and facilitates an analysis of the global dynamics of
systems with more than two ordinary differential equations.
The model is able to trace the evolution of capital and fish stock
in good agreement with observed patterns, and shows that
over-capitalization is unavoidable in unregulated fisheries.
\end{abstract}

\section{Introduction}

Following sustained interest from policy makers, recent years have
seen a number of bio-economic models examining the effects of
commercial fisheries on marine resources.
Even though over-fishing has been a fact since historical times
\citep{Jackson.2001}, the problem has gained a new quality
due to the industrialization of fisheries \citep{FAO.2004}. 
In particular the latter has reduced fish biomass by 80\% within 15 years of 
exploitation \citep{Myers.2003}. In this
context, the impact of capital accumulation has been a major issue in
fisheries economics over the last two decades
\citep{Clark.1979,McKelvey.1985,Boyce.1995,Jorgensen.1997,Munro.1999,Pauly.2002}.
In these contributions, commercial fisheries is portrayed as a system in which
a biological stock and a capital stock interact dynamically. As the
capital stock is highly specialized and cannot readily be converted to
other uses, investment decisions are characterized by
irreversibility, which is assumed as a major cause of over-fishing.

The previous literature has treated capital accumulation in various
settings. \citet{Clark.1979}
study the optimal management strategy for a renewable resource
with irreversible investment, assuming that marginal investment costs
are constant. The latter assumption is abandoned by \citet{Boyce.1995} on
the grounds that constant investment costs imply
an immediate jump in the capital stock, which is then followed by a
period of decline in both the capital and fish stock. 
Contrary, observed patterns of capital accumulation
are characterized by an initial phase of continuous growth of fleet size. 
Assuming
increasing marginal investment costs leads to better agreement with
observations, but makes the model more complicated. Considerations of
tractability therefore lead \citet{Boyce.1995} to assume that
harvest productivity is independent of the size of
the biological stock. 

In contrast to these optimal-exploitation models, approaches which
study capital accumulation in more realistic settings
are rare. An exception is
\citet{McKelvey.1985,McKelvey.1986} who
examines an open-access fishery with irreversible investment under
both perfect and imperfect competition.
But the increased realism
of these models comes at a cost
in that marginal investment costs and
harvesting productivity are kept constant
in  the analysis of out-of-equilibrium behaviour
due to serious analytical difficulties.
In general, the variety of modelling strategies pursued in the
literature thus
reflects the tension between realism and tractability, illustrating
the need for new concepts in 
integrated modelling \citep[cf.][]{Knowler.2002,Mueller.2003}.
In order to
keep them accessible to analysis, most of the models mentioned
above disregard at least one of these difficulties, e.g. those
relating to investment costs, harvesting productivity, or industry
structure.
In many cases the difficulties restrict analysis to
equilibria 
or comparative dynamics in the neighbourhood
of an equilibrium\footnote{Even with one state and one control variable
the analysis of the comparative dynamics can become 
difficult \citep{Caputo.1989,Caputo.2003}. However, some papers
investigate the dynamic properties of trajectories more thoroughly \citep[e.g.][]{Jorgensen.1997,Scheffran.2000}.}.
But since fisheries systems tend to stay
far away from
equilibrium, e.g. when catches reside above the maximum sustainable
yield, there exists an urgent need for the analysis of the enfolding
dynamics. A common approach to this task is phase plane analysis. Although
this technique is not impossible for systems with more than two
ordinary differential equations (ODEs), 
it becomes rather difficult \citep[cf.][]{Berck.1984}.
Another technique to tackle tractability problems
might be to run an ensemble of numerical simulations or to use
methods from nonlinear analysis (e.g. bifurcation analysis or
computing domains of attraction)
rather than solving analytical models. These approaches are limited
if precise parameters and exact functional specifications are not
completely available.
In fact, as pointed out
by \citet{Clark.1999}, our understanding of bio-economic systems
is characterized by low levels of knowledge. The dynamics of fish
stocks, the economic characteristics and strategies of the fishing
industry are subject to a serious lack of information
\citep[cf.][]{Pindyck.1984,Charles.2001}.

In this paper we demonstrate a qualitative simulation technique which
complements phase space analysis and numerical simulation in 
data-poor settings when nonlinear dynamics far from equilibrium are
to be investigated. The method, developed in the field of
artifical intelligence \citep[c.f.][]{Kleer.1984,Forbus.1990,Kuipers.1994},
starts from the argument that imprecise understanding 
can be formalized and used for a variety of model-based tasks (e.g. 
identification of general dynamic properties, scenario
testing, hypothesis exploration, policy advice).
The approach allows all
possible dynamic behaviours of the system 
to be characterized and classified on the basis of purely
qualitative relationships (i.e. in the absence of quantitative information).
It is used to an increasing
extent in several fields \citep[economics and finance]{Farley.1990,Benaroch.1995},
\citep[epidemiology and genetics]{Heidtke.1998,Trelease.1999},
\citep[chemistry]{Juniora.2000}, \citep[ecology]{Guerrin.2001,Bredeweg.2003},
\citep[sustainability
science]{Kropp.2001,Petschel.2001,Kropp.2002a,Schellnhuber.2002}, but seldom in bio-economics so far. 

We use qualitative simulation to investigate the dynamics of capital
accumulation in unregulated fisheries with nonlinear investment costs
and stock-dependent harvesting productivity. Our model is able to
trace the evolution of capital and fish stocks in good
agreement with observed patterns.
A main result is that the model
necessarily produces a period where
the capital stock continues to rise even after the harvest has started
to decline, i.e. the development of excess capacities is unavoidable.
In general
the paper shows that qualitative models allow us to
derive robust properties of bio-economic systems
when we have only weak knowledge at hand.

We have organized this paper as follows: in section \ref{sec2}
the basics of qualitative modelling are introduced.
Section \ref{sec3} develops an
analytical model of capital accumulation in fisheries. In
section \ref{sec4} we generalize this model to a qualitative one
to characterize its global dynamic features. 
In section \ref{sec5} we compare the results with some
development paths recently observed in
industrial fishery and draw general implications for fisheries
management and the applicability of the QDE method.
A discussion and a summary conclude
the paper.
\section{Foundations of Dynamical Qualitative Modelling}\label{sec2}
\begin{figure}[th]\centering
\includegraphics[width=0.6\linewidth]{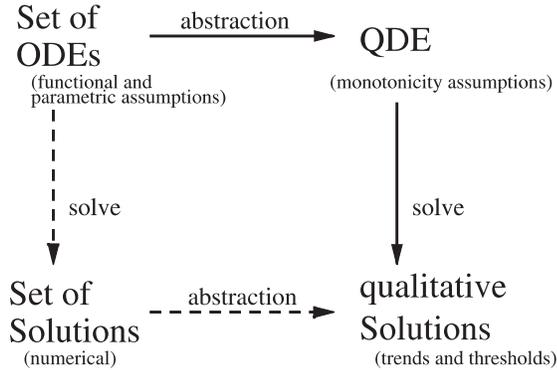}
\caption{Relationships between ordinary (ODE) and qualitative
differential equations (QDE). A reasoning process
indicated by the dashed arrows is infeasible if we want to achieve 
all numerical solutions. This becomes possible by qualitative simulation
(solid arrows) since this allows a complete computation of systems development.}
\label{fig:lab1}
\end{figure}

Qualitative differential equations (QDEs) are a prominent methodology in
qualitative modelling. For this goal \citet{Kuipers.1994} has developed the QSIM
runtime environment whose underlying concept is used for 
the analyses presented here\footnote{The original `QSIM simulator' was written in LISP and is available at \texttt{http://www.cs.utexas.edu/users/qr/QR-software.html}. For this paper we used a faster C-based version which was developed at the
Potsdam Institute for Climate Impact Research on the basis of Kuipers'
work. Together with a user guide, the source code of
the used and other related models it is provided under 
\texttt{http://www.pik-potsdam.de/\~{}kropp/compromise/qsim-bioecon.html}. \label{fn1}}.
In the following we describe the basic ideas of QDEs and
provide the necessary technical terms (written in italics). For a more detailed
introduction and a thorough overview of applications we refer to \citet{Kuipers.1994}.
The input for a modelling task is a qualitative differential equation (QDE) comprising
the following parts:
\begin{enumerate}
\item a set of continuously differentiable functions (variables) of time;
\item a quantity space for each variable, specified in terms of an 
      ordered set of symbolic {\em landmarks};
\item a set of {\em constraints} expressing the algebraic, differential or monotonic relationships between the variables. 
\end{enumerate}
A QDE can be considered as an abstract description ({\em abstraction},
cf. Fig.\,\ref{fig:lab1}) of a set of ordinary differential equations
(ODEs). The abstraction is attained in a twofold manner: (i) variables
take values from the set of symbolic {\em landmarks} or intervals
between landmarks. 
Each landmark represents a real
number, e.g. maximum
sustainable yield, of which the exact quantitative value may be
unknown. Nevertheless, it is analytically distinguished whether the catch is
above or below this threshold.
The landmark or the interval between landmarks where the value of a
variable is at a given time, is called  its {\em qualitative magnitude}. 
(ii) Monotonic relationships specified between variables, e.g. that the yield is
monotonically decreasing with a decreasing stock, are expressed by {\em constraints}.
This is an abstraction in the sense that the constraint comprises an entire
family of (linear and nonlinear) monotonic functions. The only
requirement is that these functions are smooth and that their
derivatives have certain signs.
Qualitative simulation achieves its result by performing a constraint satisfaction scheme
\citep[for a general introduction see][]{Tsang.1993,Dechter.2003}, where all
combinations of qualitative magnitudes inconsistent with the
constraints are filtered out. 
Due to continuity of the variables' values in time, the scheme additionally 
checks which admissible combinations of qualitative magnitudes
can occur after other combinations, called {\em successors}.
The guaranteed coverage theorem \citep[an in-depth discussion of this
theorem is beyond the scope of this paper, 
see][p.118]{Kuipers.1994} ensures that the algorithm
predicts an abstract description of {\em all} solutions to any ODE
described by the given QDE. 
This implies, due to the indeterminacy of analysed systems, that a QDE solution comprises not only one time development, but a set of trajectories.

In the following we illustrate these concepts considering a simple
nonlinear model, which is not meant
to solve real-world problems (it will be extended in the succeeding sections).
Consider
a natural resource stock $x$ with an associated recruitment function
$R(x)$. The industry chooses effort depending on $x$
such that there is a harvest function $h(x)$. 
The time behaviour of the stock is
expressed as
\begin{equation}
  \dot{x} = R(x) - h(x).
  \label{eq:lab1}
\end{equation}
It is assumed that $h(0)=0$, and that $h$ is strictly
increasing in $x$. The recruitment function is of
logistic type
and attains a maximum $MSY$ (maximum sustainable yield) at
$x_{MSY}$. Furthermore, $R(0)=R(Q)=0$, where $Q$ denotes
the carrying capacity of the biological system.
For $x<x_{MSY}$ the function $R$ is strictly increasing, but strictly
decreasing if $x>x_{MSY}$. 
In the first abstraction step, for each variable the following
sets of symbolic landmarks
are defined (including lower and upper bounds):
\begin{alignat}{3}
  x: & \quad 0\quad <& \quad x_{MSY}\quad < & \quad Q\quad < & x_{max},\notag \\
  h: & \quad 0\quad <& \quad MSY &\quad < \quad h_{max}, & \label{eq:lab2}\\
  R: & \quad 0\quad <& \quad R_{MSY} &\quad < \quad R_{max}. & \notag
\end{alignat}
For a given time $t$ and a (continuous) value for $x(t)$, $h(t)$ and $R(t)$ one
can determine the qualitative magnitudes with respect to the defined landmarks,
e.g. as an interval
$h(t) \in (MSY, h_{max})$, if $h > MSY$, or as a singular landmark, e.g. $h(t)=MSY$.
Qualitative simulation explicitly tracks the direction of
change of all variables in the model, which is called {\em qualitative direction}.
The arrow $\downarrow$ portrays that a variable is decreasing
(i.e. $\dot{x}(t) < 0$),
$\uparrow$ increasing ($\dot{x}(t)>0$), and a circle $\circ$ that
$\dot{x}(t)=0$ at a given time $t$.
The qualitative magnitude together with the qualitative direction of a variable
at a given time is called its
{\em qualitative value}, e.g. written as
\mbox{$\big<(MSY, h_{max}), \uparrow \big>$}. 
The qualitative values of all variables of the model calculated for a certain time is called the 
{\em qualitative state} of the system.

In the second abstraction step the constraints of the na\"ive model are
formulated as relationships
between the variables and their derivatives.
To simplify the presentation
we only introduce the most important
constraints of the model (for the whole model we refer to the
annex and to the download version;
cf. footnote\,\ref{fn1}).

\begin{figure}[ht]\centering
\includegraphics[width=0.9\linewidth]{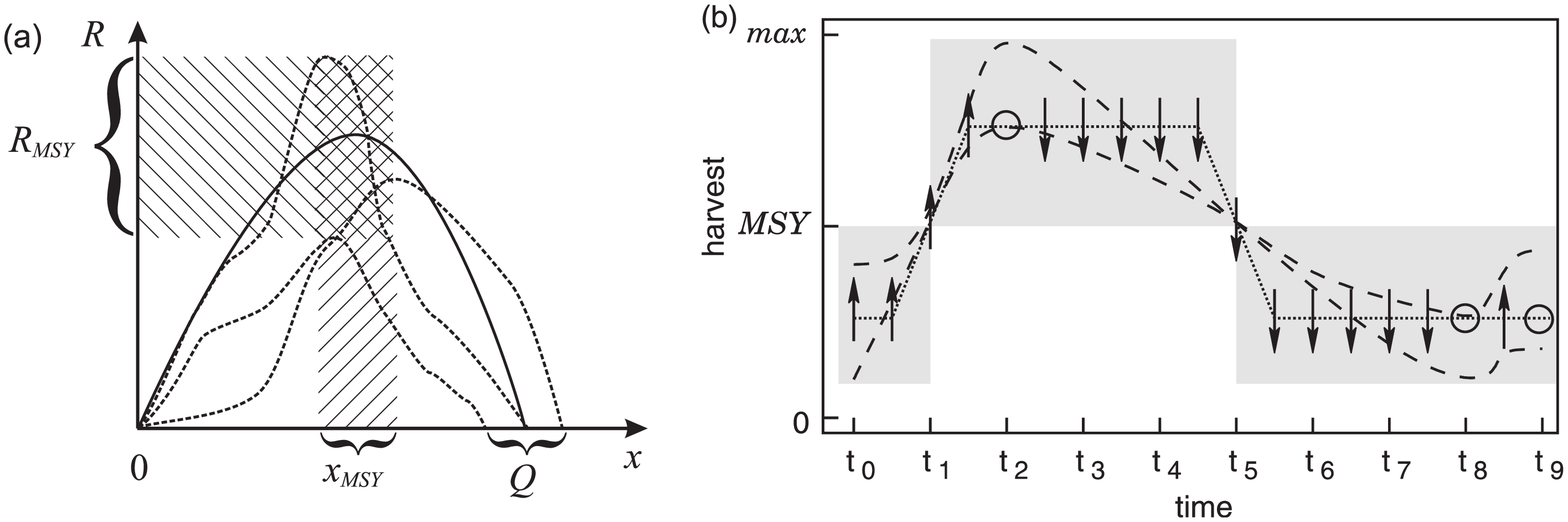}
\caption{(a) Representation of imprecise knowledge in a QDE model. 
Shown are some possible realizations of recruitment functions (dashed lines) satisfying the
  constraints\,(\ref{eq:Rrule1})--(\ref{eq:Rrule6}). The uncertainty intervals regarding
the landmarks are indicated by brackets. For $R(x_{MSY})$ all values are possible
lying in the cross-hatched area. 
(b) An example for the qualitative behaviour of harvest $h(t)$ (dotted 
line). Such a behaviour is representative for a class of (quantitative)
time
developments as indicated by the dashed lines (grey areas).}
\label{fig:lab2}
\end{figure}

Since $h_x > 0$\footnote{For the sake of
readability the first partial derivative of a function $f$ with
respect to $X$ is denoted as $f_X$ and the second derivative as
$f_{XX}$.}, 
$\dot{h}=h_x(x)\cdot \dot{x}$ and $h(0) =0$, 
using the the sign operator $sgn(\cdot)$,
\begin{equation}
 sgn(\dot{h}) = sgn(\dot{x}) \qquad \text{and} \qquad x=0 \Leftrightarrow h=0.\label{eq:hrule1} \\
\end{equation}
We define $R_{MSY}:=R(x_{MSY})$ and describe 
the relation between stock and recruitment by 
the monotonicity assumptions mentioned above by:
\begin{alignat}{4}
\text{if}\quad & x =  0   \quad    & \text{then}\quad           & R = 0, \label{eq:Rrule1} \\
               & x =  x_{MSY}      & \text{if, and only if}\quad & R = R_{MSY}, \label{eq:Rrule2} \\
\text{if}\quad & x =  Q   \quad    & \text{then}\quad           & R = 0, \label{eq:Rrule3} \\
\text{if}\quad & x < x_{MSY} \quad & \text{then}\quad &    sgn(\dot{R})=sgn(\dot{x})\quad &\text{and} \quad R < R_{MSY}, \label{eq:Rrule4}\\
\text{if}\quad & x = x_{MSY} \quad &\text{then}\quad            &    \dot{R}=0\quad &\text{and}\quad R = R_{MSY}, \label{eq:Rrule5} \\
\text{if}\quad & x > x_{MSY} \quad &\text{then}\quad            &    sgn(\dot{R})=-sgn(\dot{x})\quad & \text{and}\quad R < R_{MSY}. \label{eq:Rrule6}
\end{alignat}
The relation $\dot{x}=R-h$ implies
\begin{alignat}{5}
\text{if}\quad& R < R_{MSY}& \quad&\text{and}\quad h > MSY& \quad\text{then}\quad
& \dot{x} < 0, \label{eq:dotxrule3}
\end{alignat}
indicating that a recruitment below and a harvest above a sustainable level leads to a decreasing stock. 
It is obvious, due to the indeterminacy of the relations, that many functions exist which 
comply with these constraints (see, e.g. Fig.\,\ref{fig:lab2}a), but nonetheless
a complete qualitative simulation is possible. 
It starts with a qualitative state as
initial value at time $t_0$. 
Supposing that at the beginning the fishery is characterized by an almost
unexploited biological stock and catches above $MSY$,  
the qualitative state at $t_0$
is given by
\begin{equation}
  x(t_0) = \big<(x_{MSY}, Q), \downarrow \big>, 
  h(t_0) = \big<(MSY,h_{max}), \downarrow \big>, 
  R(t_0) = \big<(0, R_{MSY}), \uparrow \big>.
\label{eq:lab3}
\end{equation}
The qualitative magnitude of $R$ results from the qualitative magnitude of
$x$ (above $x_{MSY}$) by applying constraint\,(\ref{eq:Rrule6}). The qualitative direction of
$x$ (decreasing) is a consequence of the qualitative magnitudes of $h$ and $R$, i.e. 
rule\,(\ref{eq:dotxrule3}). Constraint\,(\ref{eq:Rrule6}) forces $R$ to
increase. Finally, $h$ decreases due to constraint\,(\ref{eq:hrule1}).
A time evolution is now initiated by these starting conditions,  
which has to respect the direction of changes shown in Eqs.\,(\ref{eq:lab3}) and the defined
constraints. For the example three possible evolutions can be distinguished:
\begin{enumerate}
\item[(i)] $x$ decreases below $x_{MSY}$, where $h$ still resides above
  $MSY$ at this time.
\item[(ii)] $h$ decreases below $MSY$, while $x$ stays above
  $x_{MSY}$ at this time.
\item[(iii)] $x$ decreases below $x_{MSY}$ and $h$ decreases below $MSY$ at
  the same time.
\end{enumerate}
These `events' (occurring at a time $t_1 > t_0$)
lead to new (and different) qualitative states. Qualitative simulation
checks which of these states comply with all constraints of
a defined QDE. Those states `surviving' this test are the successors
valid for the next time interval. For case (i), we obtain the following
qualitative state as a successor:
\begin{equation*}
  x(t_1,t_2) = \big<(0, x_{MSY}), \downarrow \big>, 
  h(t_1,t_2) = \big<(MSY, h_{max}), \downarrow \big>,
  R(t_1,t_2) = \big<(0, R_{MSY}), \downarrow \big>.
\end{equation*}
Here $t_2$ indicates the time point until the calculated state is
valid (i.e. a new `event' will occur).
The value of $x$ is the direct outcome of case (i), where
constraint\,(\ref{eq:Rrule4}) determines the value of $R$.
The magnitude of $h$ is unchanged (since otherwise, we would be in
case (iii)). Decreasing harvest is the result of constraint\,(\ref{eq:hrule1}).
By similar arguments, case (ii) leads to the successor
\begin{equation*}
  x(t_1, t_2) = \big<(x_{MSY}, Q), \downarrow \big>, 
  h(t_1, t_2) = \big<(0, MSY), \downarrow \big>, 
  R(t_1, t_2) = \big<(0, R_{MSY}), \uparrow \big>.
\end{equation*}
Case (iii) is
neglected here, because it is very unlikely that both harvest and
recruitment drop below $MSY$ exactly at the same time. In the
semantics of qualitative simulation 
the exclusion of so-called marginal cases
or other specific assumptions
can be explicitly defined (cf. annex).

\begin{figure}[t]\centering
\includegraphics[width=0.7\linewidth]{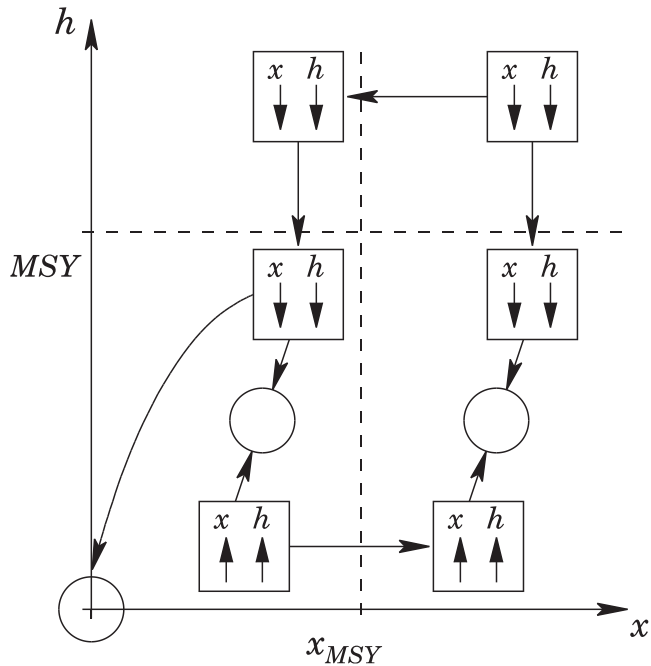}
\caption{State-transition graph (STG) for the solution set of the na\"ive model.
The arrows in the vertices (boxes) indicate the qualitative direction of the
relevant variables: $h$: harvest and $x$: stock. The position in the diagram 
refers to the qualitative magnitude, and the arrows between the boxes describe
possible changes of qualitative states in time.  Circles represent equilibrium states.
All states with a qualitative magnitude of $h>MSY$ are characterized by
a decreasing stock and harvest. Below the $MSY$ level $x$ and $h$ both
decrease or increase. One of the equilibria represents
extinction ($x=h=0$), one a steady state below $x_{MSY}$, another an
equilibrium above this landmark.}
\label{fig:lab4}
\end{figure}

In the next time steps (valid for the intervals $(t_2,t_3); (t_3,t_4);
\ldots$) qualitative simulation takes into account all successors and
performs the same consistency checks as introduced above.
The procedure is repeated
until the system attains an equilibrium, no new successor is possible
or until it enters a cycle.
A logical sequence of successors is called {\em qualitative behaviour}
and represents a qualitative dynamical trajectory.
It should be emphasized
here that a single behaviour represents a set of quantitative
development paths (including all
solutions of an ODE respecting the constraints, Fig.\,\ref{fig:lab2}b).
Since for some states more than
one successor might be possible, all states and their possible
sequences have to be displayed as a state-transition graph (STG).
Figure \ref{fig:lab4} shows the STG for the example. 
Each vertex represents a consistent qualitative state. If one state is
a successor of another, they are connected by an edge (arrow).
Each qualitative behaviour can be traced as a path along edges through
the graph. 
However,
for more complex bio-economic models the number of states typically increases
rapidly. In these cases a clustering algorithm can be applied to
provide a further structuring of the graph. For the algorithm, the
modeller has to specify a set of relevant (state) variables of the model.
All states with identical qualitative values of the relevant
variables are automatically joined to clusters. 
In the generalized state-transition graph (GSTG) each cluster is
represented by a single vertex. Its edges are inherited from the STG.
More formally, each vertex in the GSTG is an equivalence class of
vertices in the STG with respect to the qualitative value of the 
relevant variables. There is
an edge from a vertex $v$ (a class of qualitative states) to a vertex
$w$ (another class) if there exists at least one edge in
the STG where a qualitative state in $w$ is the successor of a
qualitative state in the class $v$.
This representation allows solutions of high-dimensional models 
to be displayed in a concise manner.

Comparing the QDE method with Monte Carlo simulations or
Markov chains, one has to emphasize that it is a deterministic approach.
In contrast to probabilistic methods, qualitative modelling 
determines possible transitions 
for the systems development in the logical sense.
This is quite different to techniques propagating changes on the basis
of likelihoods, e.g. by transition matrices \citep[for an example of such approaches cf.][]{Erev.1998}.
It is in general not possible to derive
probabilities in the case of multiple successors of a single state,
since these also depend on qualitative states visited before (the
Markov property is violated). Furthermore, due to various
uncertainties, it is often  
impossible to measure sufficient frequency records.
A QDE is a compact representation of a large family of ODEs, 
and the simulation result includes all their solutions.
The result is rather general in nature, since no parameters need to be
specified. Therefore, the result of qualitative simulation is 
not exactly the same as of common
phase-plane analysis.
There, multiple cases due to the number of intersections
of the main isoclines (where the derivatives of some state variables
vanish) have to be specified \citep[as in, e.g.][]{Cropper.1979} --
depending on the concrete parameterization, which is not known
to the modeller in our case.
For the same reasons, 
different cases regarding stability of equilibria, bifurcations
and other phenomena studied in
nonlinear analysis are contained in one qualitative model.
In the STG
one circle may represent multiple
equilibria of different types. If the modeller wants to exclude or
distinguish some cases, additional 
equilibria must be made explicit by introducing new landmarks and
linking them with appropriate constraints.
\section{A Fishery Model with Capital Accumulation}\label{sec3}
In this section we introduce a bio-economic model
describing a situation where $N$ identical and profit-maximizing
firms compete for an unregulated
resource, i.e. a marine fish stock with the size $x$.
Because the harvesting technology and the associated cost and profit
functions have been formulated in several ways in the previous
literature\,\citep{Clark.1979,McKelvey.1985,Boyce.1995,Amundsen.1995},
we start by deriving a generic formulation
based on standard production economics.
Assume that any harvesting
requires variable inputs (labour, fuel, material), jointly referred to
as effort $e$, and fixed inputs $k$ (capital, e.g. ships and gear).
The productivity of these inputs depends on the biological stock $x$.
Thus, a production function $f(e,x,k)$ can be defined, that determines
the resulting harvest $h \ge 0$.
For its partial derivatives
the properties $f_e, f_x, f_k > 0$, $f_{ee},f_{kk},f_{xx}<0$,
$f_{ex}, f_{ek}, f_{xk}> 0$ are imposed.
The first two sets of inequalities describe the standard properties of
positive, decreasing marginal productivity. The third set of inequalities
means that the marginal product of the variable input
decreases with decreasing fish stock, but is raised by capital
accumulation. The derivative of harvest with respect to the fish stock
also increases in capital, because certain attributes of capital
enhance the accessibility of the fish stock (improved fishing gear
and technology, increased horsepower of boats, etc.). 

Since $x$ and $k$ are given when a firm undertakes efforts $e$ to obtain a
chosen $h$, the resulting
variable costs  $v$ depend on $h,x$ and $k$. 
The function $v(h,x,k)$ can be determined
by solving $h=f(e,x,k)$ for $e$ and multiplying it with
a renumeration rate
$w$ which is assumed to be fixed.
Due to the assumptions made for $f$,
the implicit function theorem implies that
\begin{align}
 v_h > 0, v_x , v_k &< 0, \nonumber \\
 v_{hh},v_{kk},v_{xx},v_{kx} &> 0, \label{eq:pardrv} \\
 v_{hx},v_{hk} &< 0. \nonumber 
\end{align}
Additionally, we assume that the Hessian of $v$ is positive
definite, which is no contradiction to the above inequalities. This
guarantees usual convexity properties as needed later on.
It should be noted that the marginal harvesting costs $v_h$ decrease
in both the fish stock and the capital stock. This assumption differs
from previous approaches where capital only sets an upper limit for the
harvest, such that an increase in capital equipment
improves the productivity of the variable inputs, and 
capital accumulation may offset
the negative effect of a declining fish stock on harvesting costs.

We now turn to the dynamics of the economic and biological stock.
The regeneration of the resource is
given by a concave recruitment function
$R(x)$, for which the assumptions made in section \ref{sec2} are
valid. We define
\begin{equation}
  \dot{x} = R(x) - (h + h')
\label{eq:lab7}
\end{equation}
as the equation of motion,
where $h$ denotes the harvest of a firm under consideration
and $h'$ that of all the others.
Each firm's capital stock is described by
\begin{equation}
  \dot{k} = I - \delta k,
\label{eq:lab6}
\end{equation}
where $I \geq 0$ represents (irreversible) investment and
$\delta$ the depreciation rate which is assumed to be constant.
Investment costs are expressed by a strictly convex
increasing function $c(I)$. 
The convexity reflects 
inelastic supply of highly specialized equipment and rising adjustment 
costs for higher investment.
The
demand for fish is described by the downward sloping inverse
demand function $p(h+h')$.

In the following the decision of each firm on $h$ and $I$ has to be
determined.
If each fishing company acts in an economically rational way, it
chooses an
investment and harvest plan that
maximizes the discounted profit
given by
\begin{equation}
  \Pi:=\int_J e^{-rt} \bigg( p(h+h')h - v(h,x,k) - c(I) \bigg) dt
\label{eq:lab8}
\end{equation}
subject to Eq.\,(\ref{eq:lab7}) and Eq.\,(\ref{eq:lab6}).
Here, $r$ denotes a constant discount rate and $J=[0,T]$ a
planning interval.

The
optimization problem can be solved by a sufficiency theorem of
Mangasarian\,(\citeyear{Mangasarian.1966}).
By introducing $\lambda$ and $\mu$ as
costate variables for $x$ and $k$, the current-value Lagrangian
is given by
\begin{equation}
L = p(h+h')h - v(h,x,k) - c(I) +\lambda \big(R(x) - (h+h')\big)
    +\mu\big(I - \delta k\big) + \sigma_1 h + \sigma_2 I + \sigma_3 x
    + \sigma_4 k.
\label{eq:lab9}
\end{equation}
The four slack variables appear due to the constraints $I,h,x,k
\ge 0$. Since $L$ is concave in $(x,k)$, an interior solution to 
the first-order conditions $L_h=L_I=0$ 
and the costate
conditions $L_x=r\lambda-\dot{\lambda}$ and $L_k=r\mu-\dot{\mu}$
(where $\sigma_1=\sigma_2=\sigma_3=\sigma_4=0$) with
$\lambda, \mu \ge 0$ and  $\lambda(T)=\mu(T)=0$ is an optimal path. 
We assume that marginal investment costs become small for a low
investment level, i.e. $c_I(0)=0$, avoiding that there is a negative
solution of $L_I=-c_I(I)+\mu=0$ since $c_{II}>0$.
It is not as easy to guarantee a non-negative solution of $L_h=0$,
which depends on the relation between prices and
variable costs.
However, we do not investigate the optimal path in detail since it
may be unrealistic because
firms are most likely to ignore the effect of their 
harvesting decision on future stocks \citep{Harris.1998,Banks.1999,Hatcher.2000}.
In other words, they disregard Eq.\,(\ref{eq:lab7}) in their optimization
procedure,
partly because of a lack of knowledge on the recruitment function, and
partly because they consider their own influence on
the fish stock to be negligible. Moreover, they tend to assume that
other firms behave in the same way.
Thus, we suppose that
the shadow price for the biological stock is neglected by the
individual firms. We further assume that the number of firms is
constant and investment takes place in the form of increasing fishing
power or number of vessels per firm.
From this perspective it is consistent to set
$\lambda \equiv 0$ in the 
fourth summand of the current-value Hamiltonian
Eq.\,(\ref{eq:lab9}).
As a consequence, the costate condition for $\lambda$ is ignored and
corner solutions of $H_h=0$ become irrelevant.
The harvest decision is myopic in contrast to the investment decision.
Utilizing the (constant) inverse price elasticity of demand
$\epsilon < 1$, one obtains the following equations: 
\begin{align}
 L_h  &= \bigg(1-\epsilon\frac{h}{h+h'}\bigg) p(h+h') - v_h = 0,
  \label{eq:lab10a} \\
 L_I  &= - c_I(I) + \mu = 0, \label{eq:lab10b} \\
 L_k  &= - v_k - \mu \delta = r \mu -\dot{\mu}. \label{eq:lab10d}
\end{align}
According to Eq.\,(\ref{eq:lab10b}) the costate variable on capital equals the
marginal cost of investment. If the latter were constant, as assumed
in several previous models, Eq.\,(\ref{eq:lab10d}) would therefore boil down
to the usual condition that the user cost of capital, $(r+\delta) c_I$,
should just be balanced by the induced reduction in variable costs,
$-v_k$. 
In our model with increasing marginal investment costs we get a more
complicated equation. By substituting $\mu$ from Eq.\,(\ref{eq:lab10b}) and its time derivative
in Eq.\,(\ref{eq:lab10d}) one gets  $(r+\delta) c_I = - v_k + \dot{c_I}$, where
 $\dot{c_I} = c_{II} \dot{I}$.
The investment programme is characterized by the
condition that the user cost of capital should be balanced not only by
reduced variable costs, but also by the change in the purchase cost of
capital induced by a change in the level of investment. 

Because we have assumed that all firms are characterized by the same
technology and behave in the same way, one obtains $h + h'=Nh$.
Consequently, the total amount of capital and investment is
given by $N \cdot k$ and $N \cdot I$, respectively. Our model can therefore be written
in the following way:
\begin{align}
\dot{x} &= R(x) - Nh, \label{eq:ls1} \\
 v_h    &= \bigg(1-\frac{\epsilon}{N}\bigg) p(Nh), \label{eq:ls4}\\
\dot{I} &= \frac{1}{c_{II}(I)} \big( (r+\delta)c_I(I) + v_k \big),  \label{eq:ls3}\\
\dot{k} &= I - \delta k.  \label{eq:ls2}
\end{align} 
Equation (\ref{eq:ls3}) has already been interpreted
above and  Eq.\,(\ref{eq:ls4}) represents the usual equality between marginal
variable costs and marginal revenue. It should be recalled that the
marginal variable costs $v_h$ decrease in both the fish stock and the
capital stock.
Therefore, an increase in marginal costs due to a
decreasing fish stock may trigger additional investment in an effort
to keep marginal costs from rising excessively. This is crucial
for explaining why the capital stock may increase along with a
decrease in the fish stock, a phenomenon which is at the heart of what
is frequently referred to as over-capitalization. We will show in the
next section that this happens for every parameterization of
the model.

To analyse  whether over-capacities
occur or not, and whether the fish stock recovers once it
is in a critical state, we are interested in the dynamics
of the ODE model  
Eq.\,(\ref{eq:ls1})--(\ref{eq:ls2}) far from equilibrium. 
Due to the difficulties with phase-plane analysis and numerical
simulation as discussed before, we study the QDE
corresponding to the model in the next sections. 
\section{The Qualitative Model and its Solution}\label{sec4}
\subsection{Abstraction of the Analytical Model}
The abstraction procedure commences as outlined in section \ref{sec2}. 
At first, landmarks for all variables are chosen and constraints
about the functions are defined. For
the resource stock $x$, recruitment $R$ and harvest $h$ 
the same landmarks as in section \ref{sec2} are introduced. The constraints
relating $x$ to $R$ and $\dot{x}$ to $R$ and $h$
are the same as given in (\ref{eq:hrule1})--(\ref{eq:dotxrule3}) and (\ref{eq:dotxruleall}).
For capital and
investment the landmarks $k: 0 < k_{max}$ and $I: 0 < I_{max}$ are defined.
Further descriptions of the following constraints are
provided in the annex. 

Eq.\,(\ref{eq:ls4}) for the marginal variable costs $v_h$ can be solved for $h$ to yield
a harvest supply function $h(x,k)$. 
This function is increasing in both arguments, which can be shown
from the assumptions made
for the production function $f(e,x,h)$ and the inverse demand function
$p$, which implies
\begin{alignat}{3}
  \quad\text{if}\quad \dot{x}>0 &  \quad\text{and}\quad & \dot{k}>0 & &
  \quad\text{then}\quad & \dot{h}>0, \label{eq:krule} \\
  \quad\text{if}\quad \dot{x}<0 &  \quad\text{and}\quad & \dot{k}<0 & &
  \quad\text{then}\quad & \dot{h}<0.\nonumber
 \end{alignat}
The constraints following from Eq.\,(\ref{eq:pardrv}) for $v_k$ are given by
(\ref{eq:vk}), those from Eqs.\,(\ref{eq:ls3}) and (\ref{eq:ls2}) 
by (\ref{eq:irule})--(\ref{eq:iirule}) (cf. annex). For
technical reasons  $v_k$ is replaced by $-v_k: -v_{k_{max}} < 0$. 
This implies that $-v_k$ can be
expressed by a function which is strictly monotonically increasing in
$h$ and strictly monotonically decreasing
in $x$ and in $k$ (cf. Eq.\,(\ref{eq:pardrv})).
Further on, we assume that $c_{II}, r$ and $\delta$ are constant. Since
$c_I(0)=0$, it can be shown that $I$ has always the
same sign and qualitative direction as the expression
$\frac{r+\delta}{c_{II}}c_I(I)$.
Therefore, this term of Eq.\,(\ref{eq:ls3}) 
can be simplified to $I$ qualitatively. For the same reason $\delta \cdot k$ can be
replaced by $k$ (in Eq.\,\ref{eq:ls3}), and $N \cdot h$ by $h$ (in Eq.\,\ref{eq:ls1}).
Thus, the abstraction of the model and the associated constraints
as given by Eqs.\,(\ref{eq:ls1})--(\ref{eq:ls2}) 
can be expressed in the following relational form, where
the right hand side represents the typical notation developed
by \citet[cf. annex for further details]{Kuipers.1994}:
\begin{alignat}{3}
\dot{x} + h &= R   &\equiv\quad& \texttt{((add dx h R)(0 0 0)(0 MSY Rmsy))}\nonumber \\ 
R(x)&=R            &\equiv\quad& \texttt{((U- x R)(0 0)(xmsy Rmsy)(Q 0))}\nonumber\\
h(x,k) &= h        &\equiv\quad& \texttt{(((M++) x k h))}\label{math:lab3} \\
\dot{k} + k &= I   &\equiv\quad& \texttt{((add dk k I))}\nonumber \\
\dot{I} + (-v_k) &= I &\equiv\quad& \texttt{((add dI mvk I))}\nonumber \\
f(h,x,k)&=-v_k &\quad\equiv\quad& \texttt{(((M-++) h x k mvk))}\nonumber
\end{alignat}
The constraint $\texttt{add}$ is the qualitative abstraction of
quantitative addition. The $\texttt{U-}$ constraint
represents a downward bent U-shaped function of logistic type (cf. annex).
The landmarks in the brackets
are corresponding pairs/triples of argument and result values specifying
correspondences between variables, e.g. in the first equation of (\ref{math:lab3}):
if $\dot{x}=0$ and $h=MSY$, then $R=R_{MSY}$. 
\subsection{Results}
\begin{sidewaysfigure}\centering
\includegraphics[width=1\linewidth]{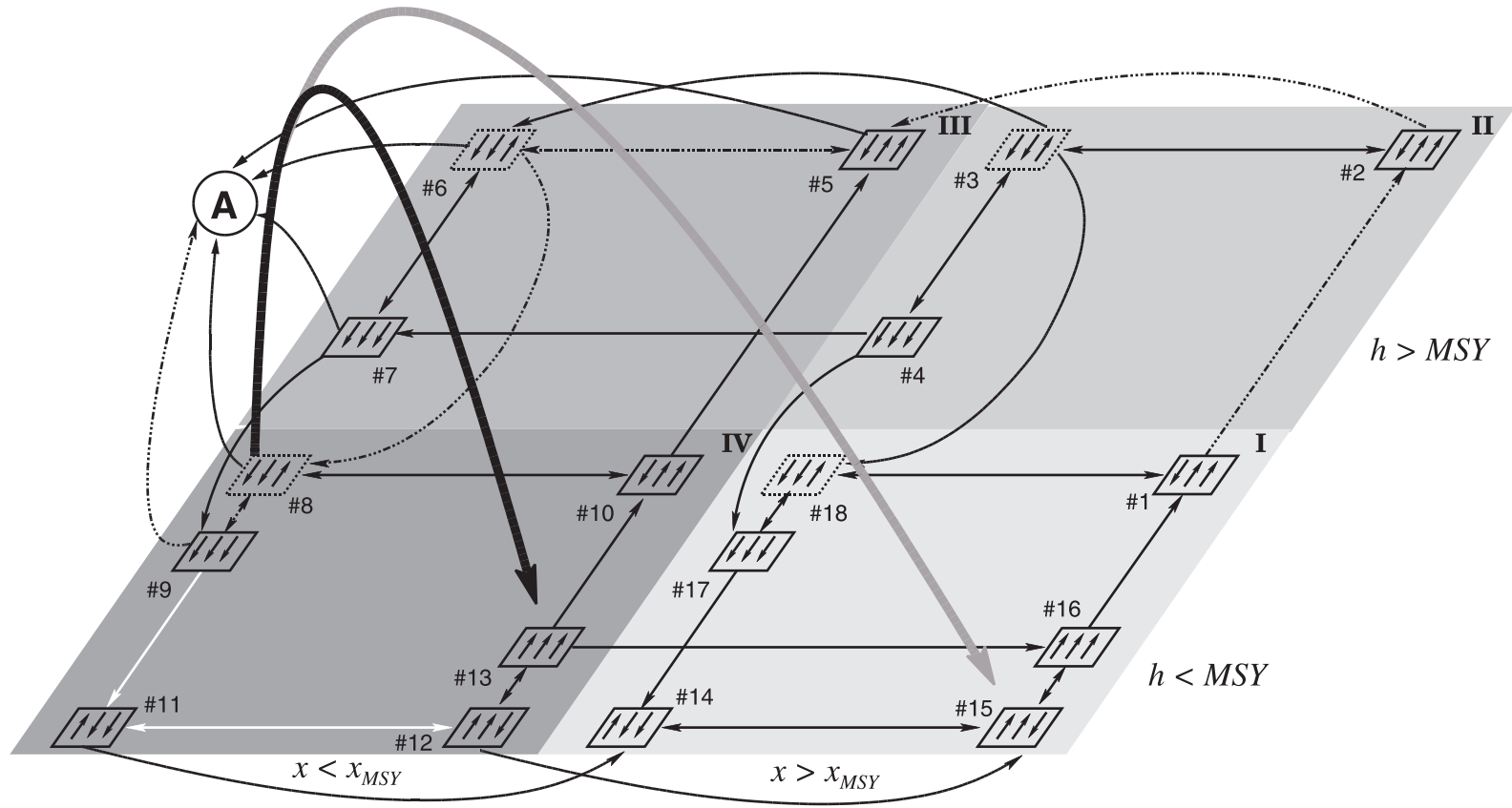}
\caption{Solutions of the qualitative model represented as GSTG. The arrows in the
  vertices label the direction of the first derivative (from left to
  right $\dot{x}, \dot{h}, \dot{k}$), the dashed boxes indicate
  situations where over-capitalization occurs. Arrows between vertices indicate
  possible transitions. The dashed and
  the white arrows indicate typical development paths (cf. text). The two
  bold arrows
  indicate  management interventions as discussed in section
  \protect\ref{sec5_2}.}
\label{fig:lab5}
\end{sidewaysfigure}

\begin{figure}[ht]\centering
\includegraphics[width=.9\linewidth]{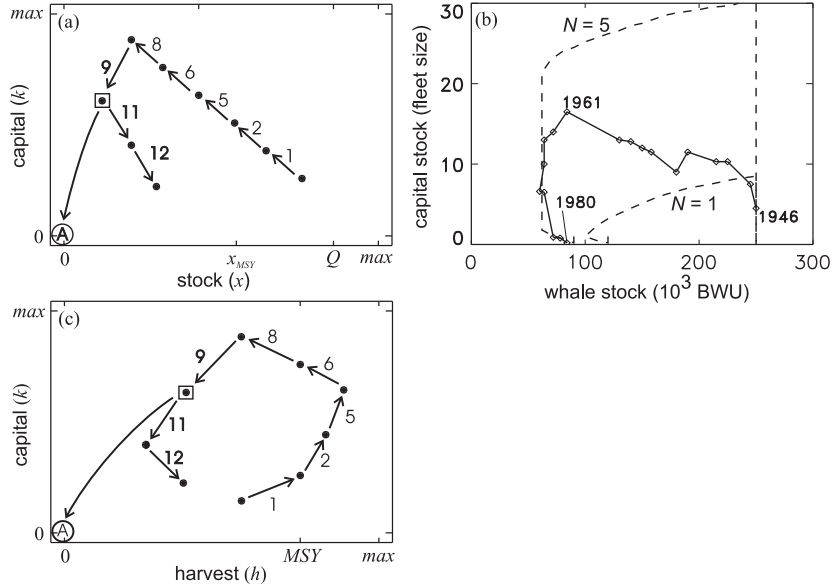}
\caption{Qualitative phase plots for two critical developments
observable in fisheries (a,c). The boxed dot refers to a 
critical branching, where the resource stock either vanishes or recovers.
The numbers correspond to the qualitative states in Fig.\,\protect\ref{fig:lab5}.
(b) displays
the situation in the blue whale hunting industry from 1946--1980
(BWU $\equiv$ blue whale units; the dashed lines refer to model
outputs generated by the \,\protect\citet{McKelvey.1986} model, and
$N$ to the number of firms in this model).}
\label{fig:lab6}
\end{figure}

The qualitative simulation of the bio-economic model (see footnote\,\ref{fn1}) 
provides 467 qualitative states. According to the methodology described in
section \ref{sec2} the potential systems developments are
structured and reorganized as a GSTG (Fig.\,\ref{fig:lab5}) with 
stock size, harvest and capital as relevant variables.
The GSTG contains 19 vertices,
where one
\mbox{{\large$\bigcirc$}\hspace*{-0.4cm}\mbox{A}\hspace*{0.1cm}},
represents a catastrophic equilibrium (where $x=0$). For each behaviour represented
in the GSTG, corresponding 
phase plots for the relevant variables are available.
Now, the question about the real-world validity of these trajectories and 
the added value for the discussion of the problems in fisheries arises. 
Our argumentation follows three directions: 
\begin{itemize}
\item[1.] In order to validate the results obtained it can be checked
  whether case studies 
  can be reconstructed by observed time behaviours. As an example
  we discuss
  the development of the blue whale fishery in the
  time period 1946-1980 (Fig.\,\ref{fig:lab6}).
\item[2.] The qualitative 
  approach determines all `dynamic patterns' which are conceivable under
  the model settings. 
  The GSTG enables us to determine general properties common to all
  time developments.
\item[3.] The GSTG has the capabilities to discuss potential
  management interventions, e.g. for scenario
  analysis or for detection of serious developments.
  In this sense the approach supplies knowledge
  for an ex-ante assessment. 
\end{itemize}
The example
of blue whale fishery (Fig.\,\ref{fig:lab6}b)
shows that the qualitative bio-economic model fits real situations
quite well. Figures \ref{fig:lab6}a and  \ref{fig:lab6}c display the associated phase
plots of the qualitative model, indicating a 
development path corresponding to transitions from vertex \#1 to \#12 (Fig.\,\ref{fig:lab5}, dashed-dotted and white arrows). Both the observed quantitative and the qualitative case
are characterized by an initial expansion phase where the whale stock
declines while the capital stock increases. This is followed by a period in
which the capital stock declines rapidly, while the whale stock is
still declining. Finally, the capital decreases further, whereas the stock tardily
recovers. This is completely different to several models 
(e.g. Clark et al. 1979, McKelvey 1986, cf. Fig.\,\ref{fig:lab6}b) which, 
due to linearity assumptions, show
an initial and a final jump of the capital stock. 

Since the GSTG contains many paths, we may wonder if every fishery
showing an intial expansion phase can be reconstructed by the model.
However, in the following we identify further structural properties of
the graph -- the model can only explain real-world fisheries which 
reflect these patterns.
During the expansion phase the
stock decreases
while capital is still increasing.
In the course of further evolution 
the effect of declining stocks on harvesting costs becomes 
so important that it cannot compensated any longer by investment. Then
the capital stock begins to decrease, i.e. net investment $I-\delta k$
becomes negative.
At this stage it is possible (Fig.\,\ref{fig:lab6}a) 
that the fish stock approaches zero (a discussion of such branching points is
provided in section \ref{sec5_1}.).
However, it is also possible that the decline of the fish stock is
reversed. In contrast to models like those of \citet{Clark.1979} and \citet{McKelvey.1986},
in this case fisheries do not need to attain an equilibrium with $x>0$, but can
exhibit a further 
boom-and-bust cycle through all quadrants. It is not even safeguarded
that the system converges to an equilibrium after repeated cycles.

We can also show that every fishery
described by the model necessarily undergoes a phase of over-capitalization.
This is related to the fact that via the simulation 
irreversible transitions are identified, i.e. transitions between qualitative states which are possible
only in one direction (cf. GSTG, e.g., from quadrant II to quadrant III). 
The proof of this proposition is as follows:
decreasing catches and increasing capital stock occur simultaneously,
whenever the system approaches the critical vertices \#3, \#6, \#8 or
\#18. If the system starts, for example, at vertex \#1
(relatively undisturbed stock) and does not directly shift to vertex
\#18, the only way to avoid vertex \#3 is to
change from vertex \#2 to \#5. However, at \#5 the only 
way for a further development (without collapse of the stock) is via vertex \#6. 
Thus, we always approach
at least one of the critical vertices and in this sense
over-capitalization is an unavoidable system property. This property is
rooted in the fact that $v_{hx}<0$ and $v_{hk}<0$, i.e. that the harvest
supply function increases in $x$ 
and $k$ (via Eq.\,(\ref{eq:ls4})). As long as
we observe increasing harvest although the fish stock is reduced, net
investment must be positive to compensate losses from increasing
marginal costs. Therefore, $k$ cannot start to decrease before $h$.
\section{Discussion and Policy Implications}\label{sec5}
Discussing the results and implications for policy actions,
a variety of conclusions can be drawn.
The model is further
validated by showing that 
mistakes arising from temporary management measures
can be reconstructed by the model.
\subsection{Why Management Fails}\label{sec5_2}
Public decision makers may respond rather late to an emerging crisis
and, in addition, with drastic, but temporary
interventions. Such situations can be analysed with the model as follows:
although the system does not follow the dynamics of an unregulated
fishery during a  
period managed in this way, we can compare the qualitative state of the system before and
after this time frame.
Two such interventions
are represented by the bold arrows in  Fig.\,\ref{fig:lab5} starting
at the dashed vertex \#8 and addressing the well-known problems in the cod fisheries of the
North Atlantic ocean (Grand Bank and Barents Sea).
The situation in the Grand Bank region and in the Barents
Sea exhibited similarities, but the results of political interventions
in these two cases were
different. Whereas in the Canadian case the worst consequences ensued,
the Norwegian authorities were able to prevent a severe disaster.
The question arises, how and why?

The cod fiasco in the Canadian waters at the beginning of the
1970s is an example of an maladjusted strategy
leading to a severe economic and environmental
catastrophe. 
\begin{figure}[ht]\centering
\includegraphics[width=0.65\linewidth]{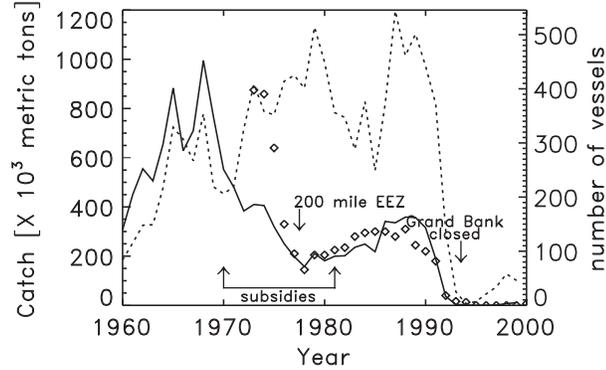}
\caption{Catch of Atlantic cod in the area
of the Northwest Atlantic Fishery Organization (Newfoundland, Grand
Banks, area 2J3KL).
The dashed line indicates the increase in number of vessels (all countries targeting cod (gadus morhua)) 
operating in this area and the
diamonds the total allowable catch 
\citep[data from][]{NAFO.2002,DFO.2002}. }
\label{fig:lab7}
\end{figure}
Anticipating an ecological disaster, the exclusive economic
zone (EEZ) was introduced in 1977 and
foreign trawlers were banned from the Grand Banks.
The government implemented catch quotas which, however, were not
always binding.
In parallel, the government responded with massive
subsidies to the fishery.
The earliest situation when such a situation occurs in the GSTG is
at vertex \#6 where the harvest is above the sustainable level
and the stock already below. The only way to avoid more serious
developments is to alter to vertex \#7 (and from there to \#9, \#11, etc.), but this
needs a reduction of the capital. Indeed the Canadians did just the opposite, i.e. 
the awareness of increasing gains due to the
ban on European trawlers led to investment in improved gear
and to more efficient fish killing\,\citep{Harris.1998}.
The situation at the end of the 1970s is represented by vertex
\#8 (Fig.\,\ref{fig:lab5}), where both stocks and catches are below a 
critical level and still decreasing while the capital continues to increase.
The reaction of the government in this situation was
counterproductive (cf. bold black arrow in Fig.\,\ref{fig:lab5}). 
It
may be that the stocks slightly recovered
(possibly indicated by an increased catch, cf. Fig.\,\ref{fig:lab7}),
but they have not reached a sustainable level (i.e. $x_{MSY}$). Thus,
when the total allowable catches (TAC) were relieved, the system was not in a safe state, 
but approached a state represented in the model results by vertex \#13.
A second period of high catch levels occurred, which led 
to the critical state \#8 again
(along the path \#13--\#10--\#8  
or via  \#13--\#10--\#5--\#6--\#8),
finally approaching \mbox{{\large$\bigcirc$}\hspace*{-0.4cm}\mbox{A}\hspace*{0.1cm}} via \# 9.
This was the end of the story, because the cod stock
has not recovered up to the present day.

To prevent these short-termed cycles,
it would be better to shift the system to vertex \#15. 
This was the case in Norway
when the country faced a similar disaster
in the Barents Sea.
Here, individual transferable quotas helped to reduce the race for fish. 
Although  there were a lot of bankruptcies in 
fisheries and many demonstrations, the politicians knew that there could
be no giving in to protests
for short-term political gain. They set up subsidies -- in contrast to the
Canadians -- to remove ships from fisheries and to diversify the
coastal economy\,\citep{Harris.1998}. Finally, they banned 
fishing from spawning grounds. This is equivalent to
a path sequence via \#8--\#9--\#11--\#14--\#15. Thus, Norway now has a better basic position
for a sustainable resource management. However, a slackening of catch permits, 
as currently discussed, may lead to the outcome that the
unregulated dynamics starts again at vertex \#15. 
We conclude from these two cases that short-term measures or
non-binding catch quotas only suspend the intrinsic problem. 
\subsection{Scenario Assessment and Critical Branchings}\label{sec5_1}
The GSTG provides various clues on how management interventions
can be implemented. 
Although some of them may not differ so much from earlier conclusions
\citep[cf. those derived from the static model
of][]{Gordon.1954}, they are here based on a comprehensive system analytical
background and a full dynamical perspective.
For example, irreversible transitions indicate
critical stages in the development of the fishery.
The fishery is in a `high-risk situation' when stock and harvest
are decreasing, but capital is still increasing and if at least one
irreversible  
transition occurred (from \#3 to \#6, or form \#4 to \#7). Critical branchings
are states characterized by multiple successors, where at least one
of them is an irreversible transition. Here, either a worsening or
mitigation is possible (e.g. \#1, \#2, \#4, \#9, or \#13).
Fishery management can use the information represented by the GSTG and
take account of the fisheries' problems in two ways: 
\begin{enumerate}
\item[(i)] How can it be avoided that
the fishery returns again to critical branching after the resource has recovered?
\item[(ii)] Under which conditions can critical branchings be prevented at
an early stage, i.e. that the system does not leave quadrants I and II?
\end{enumerate}
We can consider control measures for harvest, capital, or stock, but also - in a qualitative sense -
the time when they have to be applied at the latest.
It would also be helpful to assign likelihoods
to different transitions at these branchings. Although results of the
latter kind are problematic
in this context (cf. Sect.\,\ref{sec2}, last paragraph), the
qualitative model provides a starting point for investigations in this
direction. 
We illustrate these issues by
referring again to the example of the blue whale fishery (dashed-dotted arrows,
Fig.\,\ref{fig:lab5}). Assume that we start 
with a relatively undisturbed stock and an industry at a low level
(\#1, quadrant I, $x > x_{MSY}, h < MSY, \dot{x}<0, \dot{h}>0, \dot{k}>0$).
The first irreversible transition occurs from vertex \#1 to \#2 (quadrant II). 
Already here the catch exceeds the sustainable level ($h >
MSY$), while the capital stock is still
increasing. 
To avoid such a situation, strong harvest limitations must be introduced, i.e.
the harvest rate must decrease before $MSY$ is reached and the stock approaches $x_{MSY}$,
while over-capitalization might continue. For such a transition to
node \#18 the following scenario is conceivable:  
the harvest costs already strongly increase due to a
reduced population
although it is still above $x_{MSY}$ and/or the substitution of resource by 
capital stock has only minor effects. In reality this is observable
for less developed fisheries. For highly developed
fisheries the system evolves to vertex \#2 (quadrant II,
$x > x_{MSY},~ h> MSY,~ \dot{x}<0,~ \dot{h}>0,~ \dot{k}>0$,
cf. Fig.\,\ref{fig:lab6})  
representing fisheries having the potential to increase harvest and capital (as,
for instance, the blue whale fisheries between 1946-1960).
A safe development under these conditions
can only be guaranteed if a transition to vertex \#5 or \#6 is avoided. 
At vertex \#2 this is possible via \#3, where we are still in
quadrant II, 
but the situation is economically worse than at vertex \#2, because 
in addition to declining fish stocks, harvest is also decreasing.
Sustainable limits can only be reached by massive harvest limitations
(evolution via \#4--\#17).
If these actions are not implemented, the transition to \#5 or \#6 is
inescapable and fisheries necessarily enter to 
a situation
in which the industry and the fish stock are likely to be ruined
\mbox{{\large$\bigcirc$}\hspace*{-0.4cm}\mbox{A}\hspace*{0.1cm}}.
The further development 
depends on how early net investment
decreases, i.e. in
the phase when quadrant III is reached, effort controls are more
important than catch controls. 
In contrast in quadrant IV, successful recovery is determined by the
regeneration rate of the fish 
stock and the speed of harvest reduction.
Since policy actions often have the tendency to come
into play too late, over-capitalization and therefore over-exploitation is 
a permanent risk. Qualitative simulation
shows that at these vertices the collapse of the fishery can be prevented
if, due to a rapid capital reduction, the stock
already recovers (indicated by the white arrows,
Fig.\,\ref{fig:lab5}). In general the simulation results 
indicate that fisheries under the settings of the model
are in a state of perpetual risk, because the high risk 
states (\#5--\#9) are likely to occur repeatedly
in every boom-and-bust cycle.

We conclude that -- 
if it is impossible to formulate a precise numerical model --
qualitative simulation has
complementary advantages compared to other methods.
First, we can shift the perspective 
from equilibria to non-equilibrium dynamics and can
unveil general dynamic patterns, i.e. intrinsic properties of 
unregulated fisheries which
hold for every functional and parametric specification of the model
and which cannot be identified by comparative dynamics in a
neighbourhood of the equilibrium. 
Examples are unavoidable overcapacities and the possibility of
cyclical behaviour.
Second, several stages of system development can be identified 
to enhance our knowledge of how and when management strategies should
be introduced. This  allows alternative scenarios to be discussed.
Third, critical branchings were identified. These can be associated
with regions in the phase space where the qualitative direction of
state variables are in a configuration which admits irreversible problematic and
positive changes. In our model,
factors for the propensity of the system to recover 
include (cf. Eqs.\,(\ref{eq:ls1})--(\ref{eq:ls2})):
(i) regeneration rate of the fish stock, (ii)
depreciation of the capital stock, (iii) marginal investment costs and (iv)
marginal variable costs with respect to capital and harvest.
Here, qualitative reasoning comes to its limits, since no numerical
estimates for critical parameters can be made.
Yet we think that the method helps to identify decisive regions
of the phase space as a starting point for 
the development of new hypotheses focusing specifically on them.
Semi-quantitative techniques combining qualitative and 
quantitative methods show promise here \citep[cf.][]{Berleant.1998}.
Thus, qualitative reasoning is appropriate whenever we are dealing
with imprecise knowledge. Given the complexity
of the systems in question it must be accepted that one may have to be content with ``soft prognoses'' only.  
\section{Conclusion}\label{sec6}
This paper addresses the global
problem of industrial unregulated fisheries and the role of capital
accumulation. Such systems are often
intrinsically complex and the
understanding of them is limited by low levels of knowledge with respect to
both biological and economic properties. 
To keep models tractable, previous analytical
approaches have relied on a variety of simplifying assumptions with
respect to investment costs, harvesting costs or industry
structure. Additionally, they often concentrated on equilibrium
analysis or on
comparative dynamics near equilibrium.
We demonstrated qualitative modelling as a new method 
to approach uncertainty and tractability problems
by applying it to
a model which improves former results by
relaxing assumptions.
The qualitative model 
describes the dynamical behaviour of a fishery without 
reference to quantitative values.
We have shown that this technique can improve our reasoning
about global properties of the dynamics of the system.

The model features (i) increasing
marginal investment costs and (ii) marginal harvesting costs
that are decreasing in
both the fish and the capital stock.
Due to the former, the build-up of fishing capacities can be modelled
more realistically. Due to the latter,
fish stock and
capital stock are (incomplete) substitutes in the production of
catch.
This proved to be the key factor in
explaining why capital will keep rising while both the fish stock and the
catch decline. The qualitative simulation reveals
that all fisheries described by the general model
necessarily undergo a phase of over-capitalization.
It also
shows other inherent and -- in the sense of political
interventions -- serious system properties, e.g.
critical branchings and potentially recurring boom-and-bust cycles.

Future work will run along several lines.
Different policy measures can be assessed by incorporating them
into the qualitative model and comparing the resulting graphs.
Also a multi-species module,
together with more detailed models of decision
making in fisheries could enhance the results. Some other efforts
are related to an integration of hard and soft knowledge in
one model approach. In particular model approaches 
comprising a module for cooperative negotiations and allows to test
which management strategies are applicable in a fishery under additional settings, e.g.
normative sustainability targets, are examined \citep[see, e.g.][]{Kropp.2004d,Kropp.2005a}.
Summing up, we feel that the results and the technique open
a promising road towards new insights in the dynamics and management
of fishery systems.
\newline\newline\noindent{\bf Acknowledgements:} This work was supported by the German Federal Ministry for Education and 
Research (BMBF) under grant number 03F0205B 
(DIWA). We wish to thank the anonymous reviewers for their helpful comments.
\appendix
\section{Annex}\label{annex}
The annex completes the qualitative constraints derived from the bio-economic model
(cf. Eqs.\,(\ref{eq:ls1})--(\ref{eq:ls2}))
and the assumptions made in section \ref{sec3}.
These constraints are translated to the specific 
model code necessary to run the qualitative simulation software.

The relationship between stock $x$ and harvest $h$ is 
defined by the relations in Eq.\,(\ref{eq:hrule1}) and expressed by
constraint (C.1a) in the subsequent table.
The recruitment is given by a U-shaped function $R(x)$, defined in
Eqs.\,(\ref{eq:Rrule1})--(\ref{eq:Rrule6}) and expressed by constraint (C.1b). 
To describe the change of fish stock $\dot{x}=R-h$,
the complete model needs more than Eq.\,(\ref{eq:dotxrule3}). The derivative
$\dot{x}$ is increasing with $R$ and decreasing with $h$. In addition, if $h$ becomes zero, $\dot{x}$ changes in the
same direction as $R$ and if $R$ vanishes, $\dot{x}$ changes in the
opposite direction to $h$. Since $R_{MSY}=MSY$, $\dot{x}=0$ if
$R=R_{MSY}$ and $h=MSY$:
\begin{alignat}{5}
\text{if}\quad& R  = 0& \quad&\text{and}\quad h = 0&  \quad\text{then}\quad
& \dot{x} = 0, \nonumber \\
\text{if}\quad& R = R_{MSY}& \quad&\text{and}\quad h = MSY&  \quad\text{then}\quad
& \dot{x} = 0, \nonumber \\ 
\text{if}\quad& R < R_{MSY}& \quad&\text{and}\quad h > MSY& \quad\text{then}\quad
& \dot{x} > 0, \label{eq:dotxruleall} \\
\text{if}\quad& \dot{R} > 0& \quad&\text{and}\quad \dot{h} < 0& \quad\text{then}\quad
& \ddot{x} > 0, \nonumber \\
\text{if}\quad& \dot{R} < 0& \quad&\text{and}\quad \dot{h} > 0& \quad\text{then}\quad
& \ddot{x} < 0. \nonumber
\end{alignat}
These expressions are equivalent to those which are encoded in the qualitative
constraint (C.1c).
The constraints for the harvest supply 
function $h(x,k)$ are already defined in Eq.\,(\ref{eq:krule}) and
expressed by constraint (C.2a). 
In addition, Eq.\,(\ref{eq:pardrv}) states that the 
negative marginal costs $-v_k$ increase in $h$ and decrease in $x$ and $k$. Thus,
\begin{equation}
\begin{split}
   \text{if} \quad \dot{h}>0  \quad \text{and} \quad & \dot{x}<0  \quad \text{and}\quad 
   \dot{k}<0  \quad \text{then} \quad  -\dot{v_k} > 0, \\
   \text{if} \quad \dot{h}<0  \quad \text{and} \quad & \dot{x}>0  \quad \text{and}\quad 
   \dot{k}>0   \quad \text{then} \quad  -\dot{v_k} < 0, \\
\end{split} 
\label{eq:vk}
\end{equation}
which is expressed by the constraint (C.2b), where $v_k$ is replaced by
$mv_k := - v_k$ for technical reasons.
Taking into account Eq.\,(\ref{eq:ls2}) it is obvious that
the change of capital stock is increasing with
$I$ and decreasing with $k$. If $k$ vanishes, $\dot{k}$ changes in the
same direction as $I$, if $I$ vanishes, $\dot{k}$ changes in the
opposite direction to $k$,  yielding
\begin{alignat}{3}
        \text{if}\quad & I  = 0    &   \quad\text{and}\quad  & k = 0       &\quad \text{then}\quad &  \dot{k} = 0,   \nonumber \\
   \quad\text{if}\quad & \dot{I}>0 &   \quad\text{and}\quad  & \dot{k}<0   &\quad \text{then}\quad &  \ddot{k} > 0, \nonumber \\
   \quad\text{if}\quad & \dot{I}<0 &   \quad\text{and}\quad  & \dot{k}>0   &\quad \text{then}\quad &  \ddot{k} < 0, \label{eq:irule} \\
   \quad\text{if}\quad & \dot{I}=0 &                         &             &\quad \text{then}\quad & sgn(\ddot{k}) = -sgn(\dot{k}),  \nonumber \\
   \quad\text{if}\quad & \dot{k}=0 &                         &             &\quad \text{then}\quad & sgn(\ddot{k}) = sgn(\dot{I}).   \nonumber 
\end{alignat}
The relationship between $\dot{k}$ and $I$ is represented by (C.3).
Finally, referring to Eq.\,(\ref{eq:ls3}), the change of investment has
the form $\dot{I}=f(I)+g(v_k)$ with
strictly monotonic increasing functions $f$ and $g$ which vanish at
zero. Thus,
\begin{alignat}{3}
&\text{if}\quad I  = 0      & \quad\text{and}\quad &  v_k = 0    & \quad\text{then}\quad  & \dot{I} = 0,\nonumber \\
&\text{if}\quad \dot{I}>0   & \quad\text{and}\quad & \dot{v_k}>0 & \quad\text{then}\quad  & \ddot{I} > 0, \nonumber \\
&\text{if}\quad \dot{I}<0   & \quad\text{and}\quad & \dot{v_k}<0 & \quad\text{then}\quad  & \ddot{I} < 0, \label{eq:iirule} \\
&\text{if}\quad \dot{I}=0   &                      &             & \quad\text{then}\quad  & sgn(\ddot{I}) = sgn(\dot{v_k}),  \nonumber \\
&\text{if}\quad \dot{v_k}=0 &                      &             & \quad\text{then}\quad  & sgn(\ddot{I}) = sgn(\dot{I}),  \nonumber 
\end{alignat}
which is encoded in the constraint (C.4). Again we change $v_k$ to
$-v_k$.
The constraints, expressed by a set of comprehensible and recurring
keywords, are summarized in the following table:
\begin{center}
\renewcommand{\arraystretch}{1.2}
\setlength\tabcolsep{3pt}
\footnotesize{
\begin{tabular}{l l l}
\hline
\multicolumn{1}{c}{Index}& \multicolumn{1}{c}{Example} & \multicolumn{1}{c}{Interpretation} \\
\hline
(C.1a)&\texttt{((M+ x h) (0 0))} &  $\exists f: h=f(x), f(0)=0, f_x(x)>0$ \\
(C.1b)&\texttt{((U- x R) (xmsy Rmsy) (0 0) (Q 0))} & $\exists f:
R=f(x), f(x_{MSY})=R_{MSY}, f(0)=0,$ \\
      && $f(Q)=0; \forall x<x_{MSY}: f_x(x)>0;$\\
      && $\forall x>x_{MSY}: f_x(x)<0.$ \\
(C.1c)& \texttt{((add dx h R) (0 0 0) (0 MSY Rmsy))}& $\dot{x}+h=R \equiv \dot{x}=R-h, 0+MSY=R_{MSY}$\\
(C.2a)&\texttt{(((M + +) k x h))} & $\exists f: h=f(x,k), f_k > 0, f_x>0$ \\
(C.2b)&\texttt{(((M - + +) h x k mvk))} & $\exists f: -v_k=f(h,x,k), f_h < 0, f_x>0, f_k>0$ \\
(C.3) &\texttt{((add dk k I) (0 0 0))} & $\dot{k}+ k = I \Leftrightarrow \dot{k}=I-k$\\
(C.4) &\texttt{((add dI mvk I) (0 0 0))} & $\dot{I}+ {(-v_k)}=I$ \\
\\
\multicolumn{3}{l}{Additional constraints for derivatives and exclusion of marginal cases are:}\\
(C.5)      &\texttt{((d/dt x dx))} & $\dot{x}=\frac{d}{dt}x$  \\
(C.6)      &\texttt{((cornot x dx) (xmsy 0))} & $\forall t: x(t) \neq x_{MSY} \vee \dot{x}(t) \neq 0$, \\
      &&  i.e. it is forbidden that $x = x_{MSY}$ and \\
      && $\dot{x}=0$ at the same time \\
\hline
\end{tabular}}
\end{center}

\end{document}